\documentclass[runningheads]{llncs}
\usepackage[T1]{fontenc}
\usepackage{graphicx}
\usepackage{multirow}
\usepackage[normalem]{ulem}
\useunder{\uline}{\ul}{}
\usepackage{graphicx}
\usepackage{epstopdf}
\usepackage{caption}
\usepackage{subcaption}
\usepackage{amsmath}
\usepackage{amssymb}
\usepackage{float}
\usepackage{marvosym}

\begin{document}
\title{Multi-intent Aware Contrastive Learning for Sequential Recommendation}
\author{Junshu Huang\inst{1} \and
Zi Long\inst{2}\textsuperscript{(\Letter)} \and
Xianghua Fu\inst{2} \and
Yin Chen\inst{2}}
\institute{Shenzhen University, China
\and Shenzhen Technology University, China, \email{longzi@sztu.edu.cn}}
\maketitle       
\begin{abstract}
Intent is a significant latent factor influencing user-item interaction sequences. Prevalent sequence recommendation models that utilize contrastive learning predominantly rely on single-intent representations to direct the training process. However, this paradigm oversimplifies real-world recommendation scenarios, attempting to encapsulate the diversity of intents within the single-intent level representation. SR models considering multi-intent information in their framework are more likely to reflect real-life recommendation scenarios accurately.
To this end, we propose a \textbf{M}ulti-intent Aware \textbf{C}ontrastive  \textbf{L}earning for Sequential \textbf{Rec}ommendation (MCLRec). It integrates an intent-aware user representation learning method to enable multi-intent recognition within interaction sequences through the spatial relationships between user and intent representations.
We further propose a multi-intent aware contrastive learning strategy to mitigate the impact of pair-wise representations with high similarity. Experimental results on widely used four datasets demonstrate the effectiveness of our method for sequential recommendation.

\end{abstract}

\section{INTRODUCTION}

Recommendation systems assist users in capturing helpful information and deliver personalized recommendations to diverse users from extensive collections of items in reality. Sequential recommendation (SR) models~\cite{sasrec,caser}, which can effectively capture similar patterns of user behavior across different user-item interaction sequences, have become the state-of-the-art recommendation systems~\cite{iclrec,dssrec,bert4rec,s3rec}. SR models encode sequences into user representations by deep neural networks and finally make accurate next-item predictions that users would be interested in. Importantly, these predictions are consistent with the underlying logic of real-world recommendation systems.

Traditional SR-based approaches~\cite{GeoSAN,carnn,hgn} focus on learning from chronological sequences. This approach enables them to capture the sequential dynamics of user-item interactions. However, they exhibit limitations in identifying inter-sequence correlations, constraining their capacity to understand intricate user behavior patterns and preferences.
Sequences that exhibit similar purchase intentions in real-world shopping contexts provide valuable reference points for enhancing recommendation accuracy when predicting the next item. While reliable and precise labeled data is lacking, recent works have demonstrated that leveraging the intent similarity across diverse users to guide contrastive self-supervised learning (SSL) tasks can enhance the performance of SR models. 
Among those methods, ICLRec~\cite{iclrec} employs an expectation-maximization (EM) framework to maximize the agreement between a view of a sequence and its corresponding single intent, whose distributions are learned from all interaction sequences. It attempts to encapsulate the complex and diverse intents in real-world interaction sequences with a single-intent representation.

\begin{figure}[t]
\centering
\includegraphics[width=\textwidth]{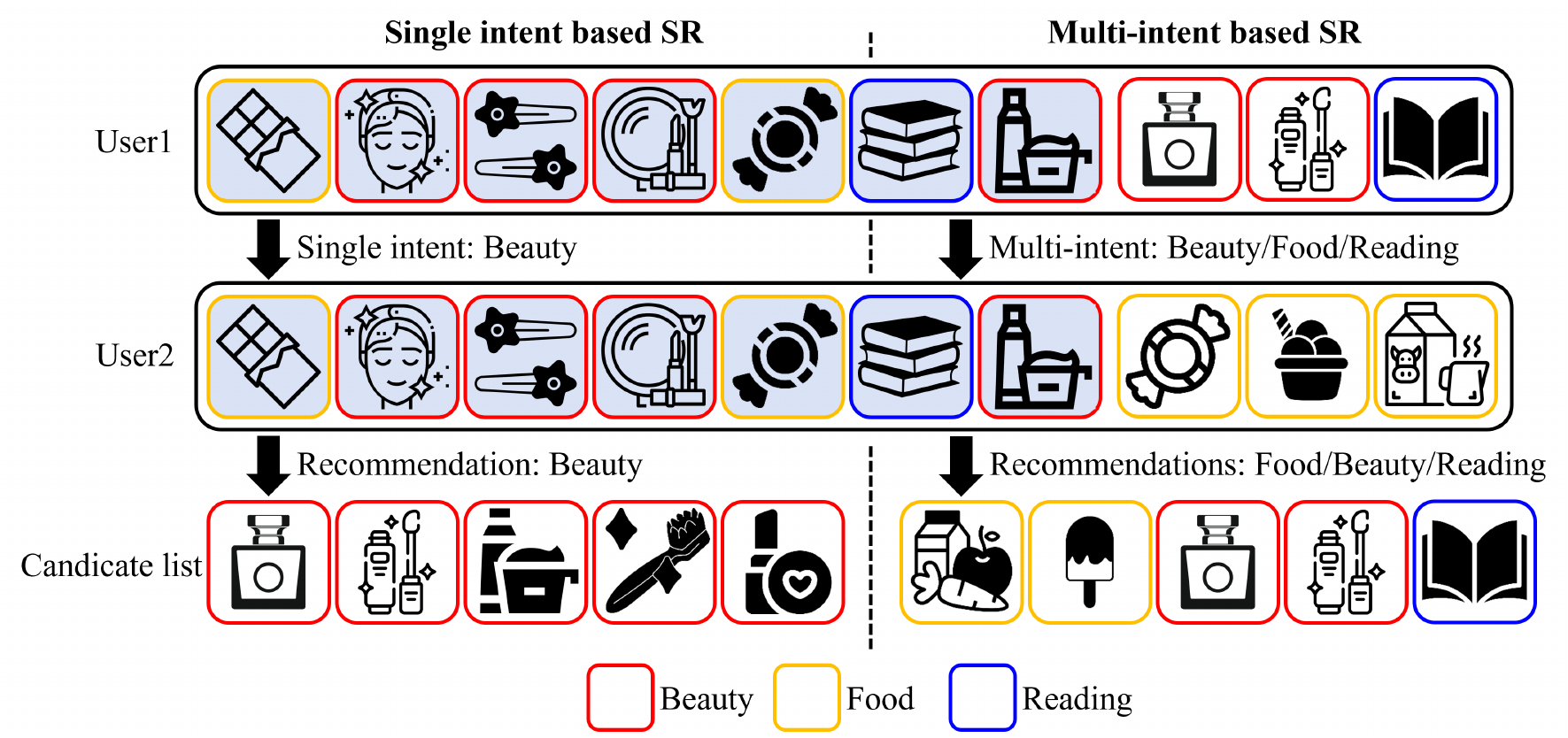}
\caption{The figure demonstrates the variation in candidate item propensity when the training of SR models is guided by single-intent or multi-intent information. Items in the sequence of User2 that are identical to those of user1 have been highlighted with a blue background.}
\label{fig:example}
\end{figure}

However, representing user-item interaction sequences with a single intent inevitably leads to losing multi-intent information. The essence of Sequence Recommendation (SR) models lies in learning sequential patterns from the training dataset to predict the next item in the testing dataset's sequences. Each user-item interaction sequence contains distinct intentions, yet models considering single-intent capture only the primary intent of each user, resulting in a loss of multi-intent information.
Consider the example illustrated in Fig. \ref{fig:example}. 
User1 and User2 have a portion of the same interaction pattern. For instance, both have engaged with face masks and hair clips in the 'Beauty' category and chocolates and candies in the 'Food' category. However, there are also some distinct interactions, such as the presence of perfume in User1's sequence, which is absent in User2's.
The SR model on the left infers the 'Beauty' intent by extracting single-intent information from User1, tending to recommendations of items from the 'Beauty' category to a similar User2. Conversely, a multi-intent aware SR model considers a range of intents such as 'Beauty', 'Food' and 'Reading', as learned from User1's behavior. The model then leverages its comprehensive understanding to offer diverse recommendations to User2, thus potentially enhancing the effectiveness of the recommendations.

SR models with multi-intent modeling are still underexplored. IOCRec~\cite{iocrec} features a global module designed to capture user preferences by disentangling the intent dimensions, thus separating global and local representations of a sequence. The sum of these two representations is utilized as the intent representation for contrastive learning (CL). This approach disentangles the multiple intents within a sequence, yet there is no information crossover in the intent dimension; in other words, a single-intent level representation is employed in the CL process. SR models incorporating multi-intent considerations into their structural design can offer a more accurate representation in actual recommendation scenarios.

To address the issues mentioned above, we propose a novel approach, 
\textbf{M}ulti-Intent Aware \textbf{C}ontrastive \textbf{L}earning for Sequential \textbf{Rec}ommendation (MCLRec), 
which utilizes multi-intent level information for model construction. 
Specifically, we apply an intent-aware user representation learning approach to infer a variety of intents within sequences and leverage the spatial relationship between user and intent representations in the latent space. 
To reduce the impact of irrelevant data, the model filters out a given number of main intents to enhance the quality of the intent-aware user representation learning. Then, we propose a multi-intent aware contrastive learning strategy, which aims to mitigate the influence of the pair-wise representations with similar multiple intents on learning, thereby improving the model's performance. We summarize the contributions of this work below:
\begin{itemize}
\item MCLRec learns intent-aware user representations in the latent space from a multi-intent perspective for user-item interaction sequences.
\item We propose a multi-intent aware contrastive learning strategy to mitigate the impact of the pair-wise representations with high similarity in their representations.
\item Experimental results on four datasets verify the effectiveness of our proposed method.
\end{itemize}

\section{RELATED WORK}

\subsection{Sequential Recommendation}

Sequential recommendation (SR) aims to disentangle users’ interest according to historical interactions, which has been widely researched~\cite{SURGE,fang2020deep,MTAM,SINE,MARank}.
Early works on SR usually extract sequential patterns based on the Markov Chain (MC) assumption. 
FPMC~\cite{FPMC} fuses sequential patterns and users’ general interests, combining a first-order MC and Matrix Factorization.
Fossil~\cite{fossil} fuses similarity-based models with a high-order MC to tackle data sparsity issues for clarity.
Recent models have begun to integrate deep neural networks into SR, such as Recurrent Neural Networks (RNN)-based~\cite{donkers2017sequential,GRU4Rec,RRN,dream} and Convolutional Neural Networks (CNN)-based~\cite{caser,yuan2019simple,DeepCoNN} models. 
GRU4Rec~\cite{GRU4Rec} first introduces RNN in session-based recommendation trained with a ranking loss function.
Caser~\cite{caser} embeds interaction sequences into images and extracts sequential patterns with CNN.
BERT4Rec~\cite{bert4rec} leverages a deep bidirectional self-attention network to model interaction sequences, utilizing the Cloze task to capture the sequential dependency effectively.
However, the methods mentioned above for SR often struggle to address issues of data sparsity and noise effectively.

\subsection{Contrastive SSL for SR}
 
SSL has emerged as a significant trend in CV~\cite{SimCLR,MoCo,cvssl}, NLP~\cite{bert,albert,roberta} and recommendation~\cite{dhcn,sqnsac,fmfd,yao2021self}. Contrastive SSL aims to extract correlation within vast amounts of unlabeled data to enhance the capability to discern negative samples simultaneously.
S\textsuperscript{3}-Rec~\cite{s3rec} proposes four self-supervised optimization objectives to capture the interrelations between items, attributes, sequences and subsequences. 
SGL~\cite{SGLRec} adopts a multi-task framework with SSL and maximizes the agreement between different augmented views of the same node to improve node representation learning.
CL4SRec~\cite{cl4srec} maximizes the agreement between differently augmented views of the same sequence in the latent space and utilizes a multi-task framework to encode the user representation.
CoSeRec~\cite{coserec} advances CL4SRec by proposing two additional data augmentation techniques to exploit item correlations.
However, the abovementioned methods do not account for users' latent intent in applying contrastive SSL.
This can limit the model's capacity to discern the nuanced motivations driving user behavior, which is critical for tailoring recommendations that align with underlying user preferences.

\subsection{Latent Intent for Recommendation}

Many recent works have focused on learning intent representation to enhance the performance and robustness of models~\cite{comirec,IDSR,ICM-SR,ASLI,MCPRN}. 
ASLI~\cite{ASLI} leverages a temporal convolutional network alongside user side information to decode latent user intents, incorporating an attention mechanism to address the complexities of long-term and short-term item dependencies.
DSSRec~\cite{dssrec} introduces a SSL task in the latent space and designs a sequence encoder to infer and disentangle the latent intents under interaction sequences.
ICLRec~\cite{iclrec} introduces a latent intent variable to maximize the agreement between user representations through an EM framework. The intent representations are used to supervise user representations clustered by K-means.
IOCRec~\cite{iocrec} suggests a novel sequence encoder integrating global and local representations to select the primary intents.
Distinct from these works, our approach incorporates multi-intent level information within one interaction sequence when learning another sequence. This enables our model to learn multi-intent aware user representations and amplify the efficacy of contrastive learning tasks.

\section{PRELIMINARIES}

\subsection{Problem Definition}

Sequential Recommendation (SR) predicts the next item users would be interested in based on their interaction sequences. 
We denote a set of users and items as $\mathcal{U}$ and $\mathcal{V}$, respectively.
Given a user $u\in \mathcal{U}$, the user sequence is a sequence of user-item interactions 
\begin{math}
  {\mathcal{S}}^u=[{s}_{1}^{u},{s}_{2}^{u},...,{s}_{\left |{\mathcal{S}}^{u} \right |}^{u}]
\end{math}, 
where $\left | \mathcal{S}^u\right |$ is the total number of interactions and ${s}_{t}^{u}\in {\mathcal{V}}$ denotes the item that user $u$ interacts with at time step $t\in [1,\left | \mathcal{S}^u\right |]$.
The sequence ${\mathcal{S}}^{u}$ is usually truncated by maximum length $T$. 
If ${\left |{\mathcal{S}}^{u} \right |}\geqslant T$, the latest T item interactions are considered, expressed as 
\begin{math}
  {\mathcal{S}}^u=[{s}_{\left |{\mathcal{S}}^{u} \right |-T+1}^{u},{s}_{\left |{\mathcal{S}}^{u} \right |-T+2}^{u},...,{s}_{\left |{\mathcal{S}}^{u} \right |}^{u}]
\end{math}; 
otherwise, zero items are padded before ${\mathcal{S}^u}$ until ${\left |{\mathcal{S}}^{u} \right |}= T$.
For convenience, $\mathcal{S}^u$ is denoted as 
\begin{math}
  [{s}_{1}^{u},{s}_{2}^{u},...,{s}_{T}^{u}].
\end{math}
The goal of SR is to predict the next item ${s}_{T+1}^{u}$ with the highest probability of interaction, which is formulated as follows:
\begin{equation}
\label{nip}
\mathop{\arg\max}\limits_{v\in\mathcal{V}}P(s^u_{T+1}=v|{\mathcal{S}}^u).
\end{equation}

\subsection{Next Item Prediction}

The main objective of the next item prediction is to develop an encoder $f_\theta(\cdot)$ that takes interactions 
$\mathcal{S}=\left \{ \mathcal{S}^u\right \}_{u=1}^N$ 
as input and generates user representations 
$\mathcal{H}=\left \{ \mathcal{H}^u\right \}_{u=1}^N$ 
as output in a batch with $N$ users. Here, 
$\mathcal{H}^u=[\mathbf{h}_1^u,\mathbf{h}_2^u,...,\mathbf{h}_{T}^u]$ and $\mathbf{h}_{t}^u$ represents interacting item of user $u$ at step $t$.
According to Eq. (\ref{nip}), parameters $\theta$ can be optimized by maximizing the log-likelihood of the next items of $N$ sequences, as expressed by the formula:
\begin{equation}
\mathop{\arg\max}\limits_{\theta}\displaystyle\sum_{u=1}^{N}\sum_{t=2}^{T}lnP_\theta(\mathbf{s}^u_t),
\end{equation}
where $\mathbf{s}^u_t$ represents the embedding of target item $s^u_t$. To achieve this, the adapted binary cross entropy loss can be equivalently minimized, defined as:
\begin{equation}
\label{recloss}
  \mathcal{L}_{Rec}=-\displaystyle\sum_{u=1}^{N}\displaystyle\sum_{t=2}^{T}\left [ \log{\sigma \left ( \mathbf{h}_{t-1}^u\cdot \mathbf{s}_t^u\right )}+ \displaystyle\sum_{neg}\log{ \left ( 1-\sigma \left ( \mathbf{h}_{t-1}^u\cdot \mathbf{s}_{neg}^u\right ) \right )}\right ],
\end{equation}
where $\mathbf{s}_{neg}^u$ is the embedding of item never interacted with by user $u$ and $\sigma$ represents the sigmoid function.
A sampled softmax technique is adopted to reduce computational complexity, following the approach in S\textsuperscript{3}-Rec~\cite{s3rec}, where a negative item is randomly sampled for each time step in each sequence. 

\subsection{Contrastive Learning in SR}
\label{CLinSR}

By adopting the mutual information maximization principle, the contrastive learning (CL) paradigm for SR leverages correlations among different views of the same sequence, maximizing user representations' agreement and enhancing the learning process. InfoNCE, as one of the CL approaches, aims at optimizing a lower bound of mutual information~\cite{SimCLR,MoCo}. 
Given a user-item interaction sequence ${\mathcal{S}}_u$, sequence augmentation is operated to create two positive views $\alpha$ and $\beta$ denoted as ${\tilde{\mathcal{S}}^u_\alpha}$ and ${\tilde{\mathcal{S}}^u_\beta}$, that can be formulated as follows:
\begin{equation}
  {\tilde{\mathcal{S}}^u_\alpha}={g}_{\alpha}({\mathcal{S}}^u),
  {\tilde{\mathcal{S}}^u_\beta}={g}_{\beta}({\mathcal{S}}^u),
\end{equation}
where $g_\alpha$ and $g_\beta$ are randomly chosen as augmentation approaches from 'crop', 'mask', or 'reorder' like BERT4Rec~\cite{bert4rec} and CL4SRec~\cite{cl4srec}.
We usually treat two views created from the same sequence as positive pairs and, conversely, created from different sequences as negative pairs.
These views are encoded to two-dimensional ${{\tilde{\mathcal{X}}}^u_\alpha},{{\tilde{\mathcal{X}}}^u_\beta}\in {\mathbb{R}}^{T\times d}$ with the sequence encoder ${f}_{\theta }(\cdot )$, expressed as ${\tilde{\mathcal{X}}}^u_\alpha=f_{\theta}({\tilde{\mathcal{S}}}^u_\alpha)$
and ${\tilde{\mathcal{X}}}^u_\beta=f_{\theta}({\tilde{\mathcal{S}}}^u_\beta)$,
and then are concatenated into one-dimensional vectors as ${\tilde{\mathbf{x}}^u_\alpha},{\tilde{\mathbf{x}}^u_\beta}\in {\mathbb{R}}^{Td}$, where $d$ is the embed size of the encoder ${f}_{\theta }(\cdot )$.
Finally, parameters $\theta$ can be optimized by the InfoNCE loss function:
\begin{equation}
\label{clloss}
  {\mathcal{L}}_{CL}={\mathcal{L}}_{CL}\left ({\tilde{\mathbf{x}}^u_\alpha},{\tilde{\mathbf{x}}^u_\beta} \right )+{\mathcal{L}}_{CL}\left ({\tilde{\mathbf{x}}^u_\beta},{\tilde{\mathbf{x}}^u_\alpha} \right ),
\end{equation} 
and
\begin{equation}
  {\mathcal{L}}_{CL}\left ({\tilde{\mathbf{x}}^u_\alpha},{\tilde{\mathbf{x}}^u_\beta} \right )={-log}\frac{exp({\tilde{\mathbf{x}}^u_\alpha}\cdot{\tilde{\mathbf{x}}^u_\beta})}{\textstyle\sum_{neg}{exp({\tilde{\mathbf{x}}^u_\alpha}\cdot{\tilde{\mathbf{x}}^u_{neg}})}},
\end{equation}
where ${\mathbf{x}^u_{neg}}$ is a negative view representation of ${\mathbf{x}^u_\alpha}$.

\section{THE PROPOSED METHOD}

\begin{figure}[!t]
    \centering
    \begin{subfigure}[t]{\textwidth}
        \centering
        \includegraphics[width=\textwidth]{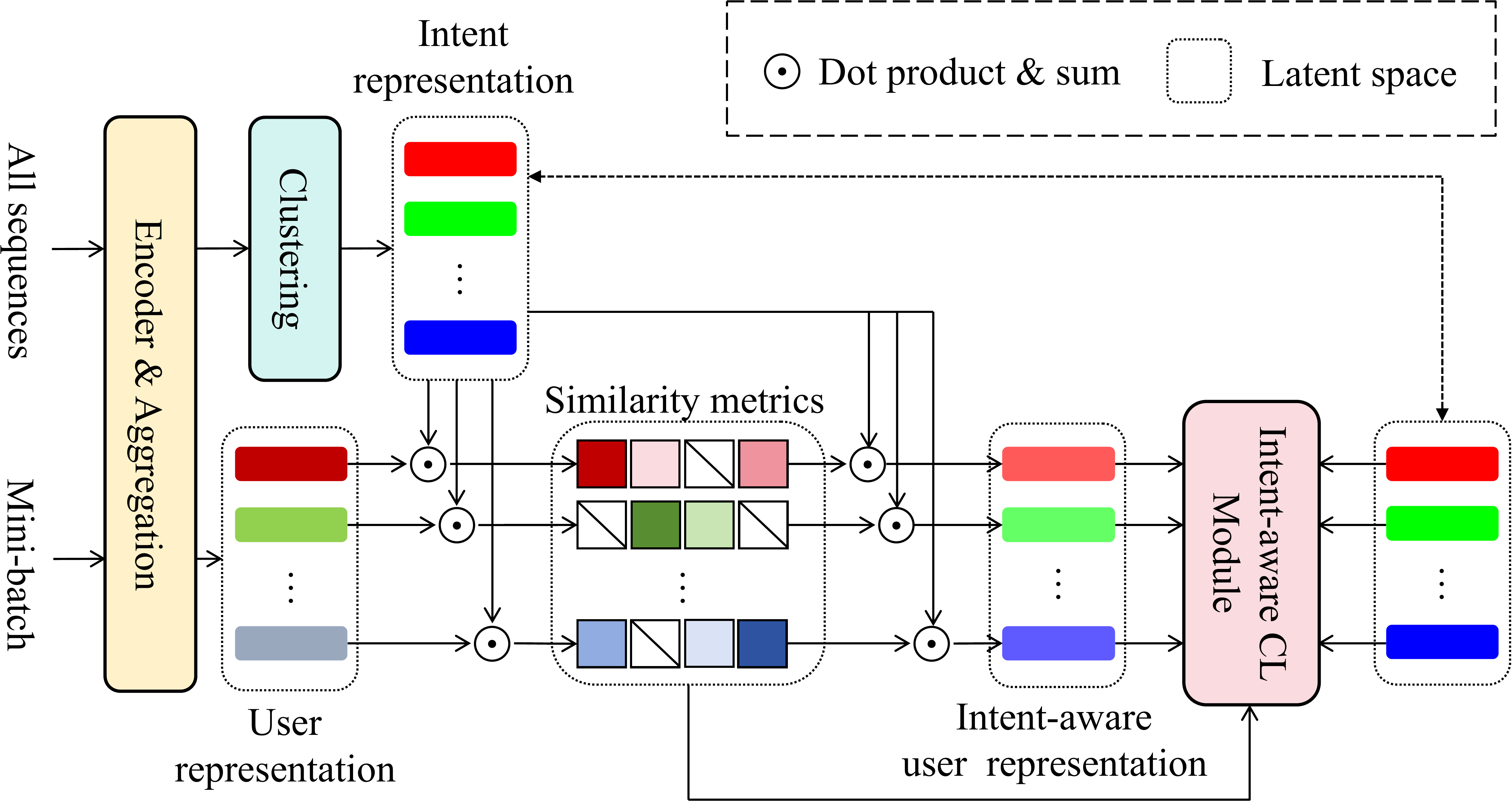}
        \caption{The proposed MCLRec.}
    \end{subfigure}
    \begin{subfigure}[t]{\textwidth}
        \centering
        \includegraphics[width=0.6\textwidth]{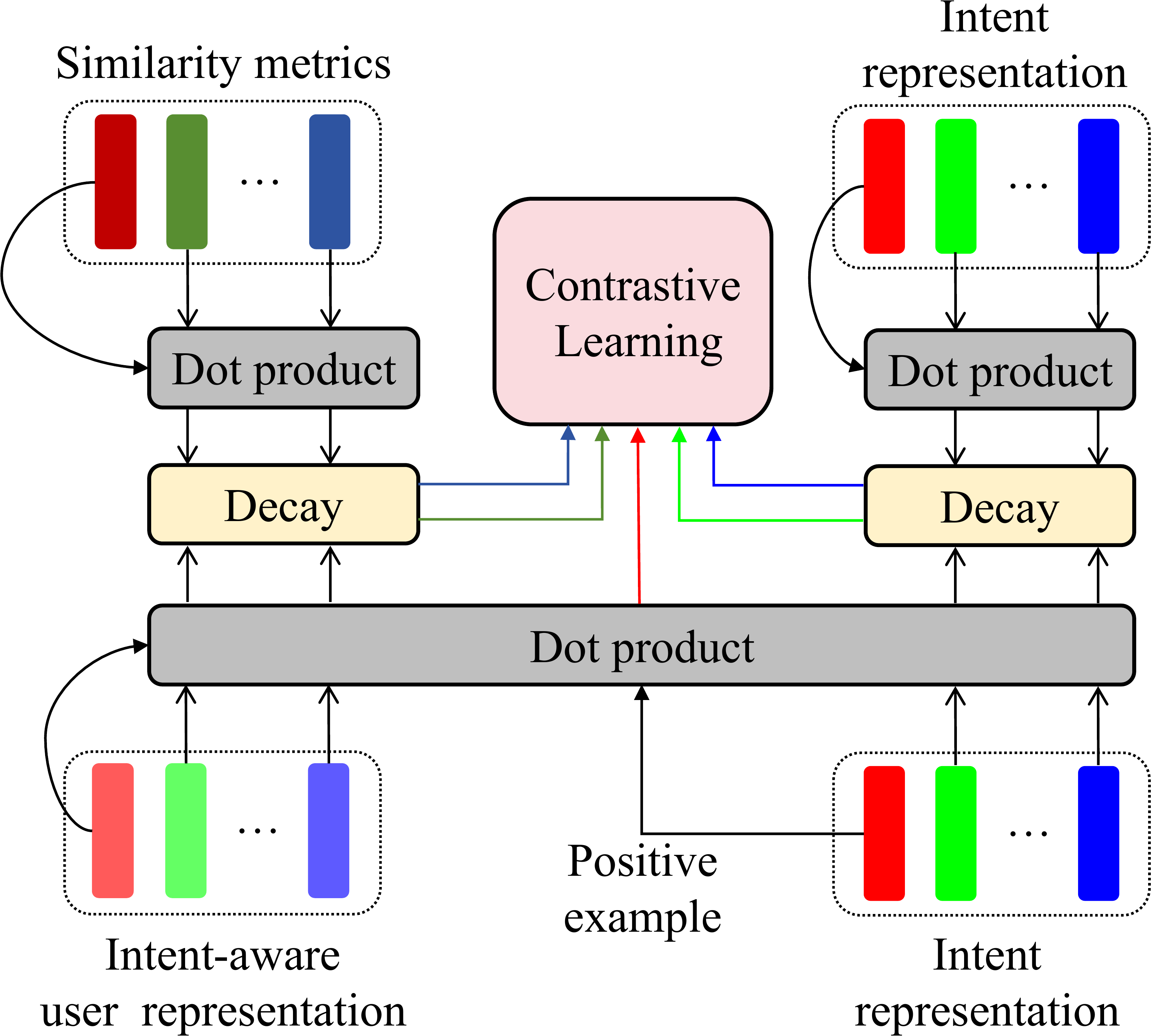}
        \caption{The process of the multi-intent aware contrastive learning task.}
    \end{subfigure}
    \caption{Overall framework.}
    \label{fig:model}
\end{figure}

Fig. \ref{fig:model} shows the framework of the proposed MCLRec. It first estimates intent representations from all interaction sequences. Subsequently, it computes similarity metrics by comparing intent representations with user representations and further obtains the intent-aware user representations after masking low correlation metrics. 
We propose a CL method to enhance the learning quality of intent-aware user representations. Instead of distinguishing between positive and negative examples in the traditional sense, we employ multi-intent aware weights, which are continuous, to quantify the similarity between samples and the learning objective by weight decay approach.
The framework is designed to leverage multi-intent level information throughout the training process.

\subsection{Intent-aware User Representation}

The intent-aware user representations are estimated through clustering by 
$\mathcal{H}^{all}=\{\mathbf{h}^u\}_{u \in \mathcal{U}}$, 
which denotes the set of all user representations. We utilize $\mathcal{H}^{all}$ to calculate a set of $K$ cluster centroids 
$\mathcal{C}=\{ \mathbf{c}^i \}_{i=1}^K$ 
as latent intent representations by a K-means clustering $\mathcal{K}(\cdot)$, expressed as $\mathcal{C}=\mathcal{K}(\mathcal{H}^{all})$. 
Each centroid $\mathbf{c}_i$ represents a cluster corresponding to a specific latent intent inferred from users' interaction patterns in $\mathcal{U}$. 

As shown in Fig. \ref{fig:model}(a), we can aggregate the user representations 
$\mathcal{X}^u_\alpha=[\mathbf{x}_{\alpha,1}^u,\mathbf{x}_{\alpha,2}^u,...,\mathbf{x}_{\alpha,T}^u]\in {\mathbb{R}}^{T\times d}$ 
mentioned in Section \ref{CLinSR} into  
$\bar{\mathbf{x}}_\alpha^u=\frac{1}{T}\textstyle\sum_{t=1}^{T}\mathbf{x}_{\alpha,t}^u$ ($\bar{\mathbf{x}}_\alpha^u \in {\mathbb{R}}^{d}$ ) 
to reduce computational complexity through mean pooling. 
Similarly, we can aggregate $\mathcal{X}^u_\beta$ into $\bar{\mathbf{x}}^u_\beta$.
For the sake of convenience in description, we denote ${\bar{\mathbf{x}}^u_\alpha}$ and ${\bar{\mathbf{x}}^u_\beta}$ collectively as ${\bar{\mathbf{x}}^u}$ to indicate the application of the same operation to both, that is, 
${\bar{\mathbf{x}}^u}=\{ {\bar{\mathbf{x}}^u_\alpha}, {\bar{\mathbf{x}}^u_\beta} \}$.

Given N user representations $\{\bar{\mathbf{x}}^u\}_{u=1}^N$ in a mini-batch, the correlation of intent $i$ to user $u$ can be calculated from $\mathcal{C}$ as follows:
\begin{equation}
  w^{u,i}=\frac{1}{\left | \bar{\mathbf{x}}^u-\mathbf{c}^i\right |},
\end{equation}
where $w^{u,i}$ means the reciprocal of Euclidean distance between user representation $\bar{\mathbf{x}}^u$ and intent representation $\mathbf{c}^i$ in the same latent space. $u,i$ are in range of $[1,...,N],[1,...,K]$, respectively.

For user $u$, we assume that there are only $R$ relevant intents $\hat{\mathcal{C}^u}=\{ \mathbf{c}^{u,k} \}_{k=1}^R$ ($\mathbf{c}^{u,k} \in \mathcal{C}$) that mainly influence user's decisions within interaction sequence ${\mathcal{S}}^u$, where $R\in \left ( 0,K\right )$ is a hyper-parameter and $\mathbf{c}^{u,k} \in \mathcal{C}$ denotes the intent representation corresponding to the $k$-th largest weight for user $u$. 
Since the remaining $K-R$ intents have little impact on user decision-making, we uniformly filter out these lesser weights and set them to constant zero\setcounter{footnote}{0}\footnote{We set the values to zero before normalization and softmax operations for smoothing the weights.}. After normalization and softmax, we have found an approach to describe the weight of diverse intents to user $u$ as follows:
\begin{equation}
\hat{w}^{u,i} = 
 \dfrac{\exp{{\omega}^{u,i}}}{\sum_{j=1}^{K}\exp{{\omega}^{u,j}}},
\end{equation}
and
\begin{equation}
{\omega}^{u,i}=
\left\{\begin{matrix}
 0,& \mathbf{c^i}\notin \hat{\mathcal{C}^u}\\ 
 w^{u,i},& \mathbf{c^i}\in \hat{\mathcal{C}^u}
\end{matrix}\right.,
\end{equation}
where $u,i$ are in range of $[1,...,N],[1,...,K]$, respectively.

The metrics 
$\bar{\mathbf{w}}^{u}=
[\bar{\mathbf{w}}^{u,1},
\bar{\mathbf{w}}^{u,2},...,
\bar{\mathbf{w}}^{u,K}]$ can customize an intent-aware user representation $\bar{\mathbf{c}}^u$ based on multiple intents for sequence ${\mathcal{S}}^u$:
\begin{equation}
\label{multi_centroids}
  \bar{\mathbf{c}}^u
  =\displaystyle\sum_{i=1}^{K} \hat{w}^{u,i}\cdot \mathbf{c}^{u,i}.
\end{equation}

\subsection{Multi-intent Aware Contrastive Learning}

We have estimated intent-aware user representation set $\bar{\mathbf{c}}^u$ for user $u$ in a mini-batch. However, directly contrastive learning is not effective enough since pair-wise representations with highly divergent multi-intent are far more valuable than those with minimal differences.
We suggest assessing the relationship between representations by employing spatial distance in the latent space as shown in Fig. \ref{fig:model}(b).

In a mini-batch, we construct a merged representation set $\mathcal{B}=\{\bar{\mathbf{x}}^u, \bar{\mathbf{c}}^u\}_{u=1}^N$ consisting of $2N$ representations for CL. 
For the user representation $\bar{\mathbf{x}}^u$, we treat $\bar{\mathbf{c}}^u$ as a learning target, and the remaining $2N-1$ representations denoted by the set $\mathcal{B}^-$. The loss function about $\bar{\mathbf{x}}^u$ is optimized according to the following formulation:
\begin{equation}
\label{mcl1}
  {\mathcal{L}}_{MCL}(\bar{\mathbf{x}}^u,\bar{\mathbf{c}}^u)=
  {-\log}\frac
  {\exp ( sim(\bar{\mathbf{x}}^u, \bar{\mathbf{c}}^u)  )}
  {\sum_{\mathbf{b} \in \mathcal{B}^-}
  {\exp \left ( sim(\bar{\mathbf{x}}^u, \mathbf{b}) \right ) }}.
 \end{equation}
Analogous to Eq. (\ref{mcl1}), the loss function with respect to $\bar{\mathbf{c}}^u$ is optimized as follows:
\begin{equation}
\label{mcl2}
  {\mathcal{L}}_{MCL}(\bar{\mathbf{c}}^u,\bar{\mathbf{x}}^u)=
  {-\log}\frac
  {\exp ( sim(\bar{\mathbf{c}}^u, \bar{\mathbf{x}}^u)  )}
  {\sum_{\mathbf{b} \in \mathcal{B}^-}
  {\exp \left ( sim(\bar{\mathbf{c}}^u, \mathbf{b}) \right ) }}.
 \end{equation}
Overall, the loss function for multi-intent aware contrastive learning tasks can be articulated as follows:
\begin{equation}
\label{mclloss}
  {\mathcal{L}}_{MCL}=
  {\mathcal{L}}_{MCL}(\bar{\mathbf{x}}^u,\bar{\mathbf{c}}^u)+
  {\mathcal{L}}_{MCL}(\bar{\mathbf{c}}^u,\bar{\mathbf{x}}^u).
 \end{equation}

In Eq. (\ref{mcl1}) and Eq. (\ref{mcl2}), the function $sim(\cdot)$ acts as a similarity metric that quantifies the agreement between two representations $\mathbf{b}^p,\mathbf{b}^q\in\mathcal{B}$, as defined below:
\begin{equation}
\label{sim}
sim(\mathbf{b}^p,\mathbf{b}^q)=
\mathbf{b}^p\cdot\mathbf{b}^q-\mathcal{D}(\mathbf{b}^p,\mathbf{b}^q),
\end{equation}
and
\begin{equation}
\label{decay}
\mathcal{D}(\mathbf{b}^p,\mathbf{b}^q)=
\begin{cases}
+\infty,  & \mathbf{b}^p=\mathbf{b}^q \\ 
\log_{2}{\dfrac{2}{1-sim_{cos}(\bar{\mathbf{w}}^p,\bar{\mathbf{w}}^q)}}, 
& \mathbf{b}^p\neq \mathbf{b}^q\ and\ \mathbf{b}^p,\mathbf{b}^q\notin  \{ \bar{\mathbf{c}}^u \}_{u=1}^N  \\ 
\log_{2}{\dfrac{2}{1-sim_{cos}( \bar{\mathbf{c}}^p, \bar{\mathbf{c}}^q )}}, & otherwise
\end{cases},
\end{equation}
where $sim_{cos}(\cdot)$ denotes the cosine similarity function. 
We introduce a decay function $\mathcal{D}(\cdot)$ to control the impact of representations $\mathbf{b}^p$ and $\mathbf{b}^q$ in CL according to their similarity. 
When considering solely the similarity between user representations, the decay is determined by the parameter $\bar{\mathbf{w}}^{u}$. Otherwise, it is governed by the intent-aware user representation $\bar{\mathbf{c}}^u$.
Two proximate representations in the latent space exhibit a high degree of similarity from a multi-intent perspective, necessitating a more pronounced decay. Conversely, greater distance warrants less decay. This ensures that the similarity function respects the underlying multi-intent aware framework in the latent space by mitigating the impact of pair-wise representations with high similarity.

\subsection{Multi-task Learning}

We employ a multi-task learning framework that simultaneously optimizes the main task of sequential prediction alongside three auxiliary learning objectives. In the framework, Eq. (\ref{recloss}) is to optimize the main next item prediction task, Eq. (\ref{clloss}) is to optimize the sequential contrastive learning task, and Eq. (\ref{mclloss}) is to optimize the multi-intent aware contrastive learning task. 
Following ICLRec~\cite{iclrec}, we can optimize the intent contrastive learning task as ${\mathcal{L}}_{ICL}$.
Mathematically, we jointly train the model as follows:
\begin{equation}
{\mathcal{L}}={\mathcal{L}}_{Rec}+\beta\cdot{\mathcal{L}}_{CL}+\lambda\cdot{\mathcal{L}}_{ICL}+\gamma\cdot{\mathcal{L}}_{MCL},
\end{equation}
where $\beta, \lambda$ and $\gamma$ control the strength of CL, ICL and multi-intent aware contrastive learning tasks, respectively, to be tuned.

\begin{table}[t]
\centering
\caption{Statistics of four experimented datasets.}
\begin{tabular}{c|llll}
\hline
Dataset & Beauty & \multicolumn{1}{c}{Sports} & \multicolumn{1}{c}{Toys} & \multicolumn{1}{c}{Yelp} \\ \hline
\# Users & 22,363 & 35,598 & 19,412 & 30,431 \\ 
\# Items & 12,101 & 18,357 & 11,924 & 20,033 \\ 
\# Actions & 198,502 & 296,337 & 167,597 & 316354 \\ 
\# Avg.length & 8.9 & 8.3 & 8.6 & 10.4 \\ 
Sparsity & 99.93\% & 99.95\% & 99.93\% & 99.95\% \\ \hline
\end{tabular}
\label{table1}
\end{table}

\begin{table}[!t]
\centering
\caption{Performance Comparison on HIT and NDCG Metrics. Best baseline scores are underlined; scores where our model outperforms the baseline are in bold. The last row indicates the performance increase of our model over the best baseline as a percentage.}
\label{tab:baseline}
\begin{subtable}{1\textwidth}
\centering
\subcaption{HIT}
\begin{tabular}{c||cccc|cccc}
\hline
Metric & \multicolumn{4}{c|}{HIT@5} & \multicolumn{4}{c}{HIT@10} \\ \hline 
DataSet & Beauty & Sports & Toys & Yelp & Beauty & Sports & Toys & Yelp \\ \hline \hline
BPR-MF & 0.0178 & 0.0123 & 0.0122 & 0.0131 & 0.0296 & 0.0215 & 0.0197 & 0.0246 \\ \hline
GRU4Rec & 0.0180 & 0.0162 & 0.0121 & 0.0154 & 0.0284 & 0.0258 & 0.0184 & 0.0265 \\
Caser & 0.0251 & 0.0154 & 0.0205 & 0.0164 & 0.0342 & 0.0261 & 0.0333 & 0.0274 \\
SASRec & 0.0377 & 0.0214 & 0.0429 & 0.0161 & 0.0624 & 0.0333 & 0.0652 & 0.0265 \\ \hline
BERT4Rec & 0.0360 & 0.0217 & 0.0371 & 0.0186 & 0.0601 & 0.0359 & 0.0524 & 0.0291 \\
S\textsuperscript{3}-Rec & 0.0189 & 0.0121 & 0.0456 & 0.0175 & 0.0307 & 0.0205 & 0.0689 & 0.0283 \\
CL4SRec & 0.0401 & 0.0231 & 0.0503 & 0.0218 & 0.0642 & 0.0369 & 0.0736 & 0.0354\\ \hline
DSSRec & 0.0408 & 0.0209 & 0.0447 & 0.0171 & 0.0616 & 0.0328 & 0.0671 & 0.0297 \\
ICLRec & 0.0500 & 0.0290 & {\ul 0.0597} & {\ul 0.0240} & 0.0744 & 0.0437 & {\ul 0.0834} & {\ul 0.0409} \\
IOCRec & {\ul 0.0511} & {\ul 0.0293} & 0.0542 & 0.0222 & {\ul 0.0774} & {\ul 0.0452} & 0.0804 & 0.0394 \\ \hline \hline
MCLRec & \textbf{0.0566} & \textbf{0.0308} & \textbf{0.0635} & \textbf{0.0255} & \textbf{0.0811} & \textbf{0.0465} & \textbf{0.0896} & \textbf{0.0421} \\ 
improv. & 10.76\% & 5.19\% & 6.31\% & 6.38\% & 4.78\% & 2.92\% & 7.41\% & 3.01\% \\ \hline
\end{tabular}
\label{tab:hit}
\end{subtable}
\begin{subtable}{1\textwidth}
\centering
\subcaption{NDCG}
\begin{tabular}{c||cccc|cccc}
\hline
Metric & \multicolumn{4}{c|}{NDCG@5} & \multicolumn{4}{c}{NDCG@10} \\ \hline
DataSet & Beauty & Sports & Toys & Yelp & Beauty & Sports & Toys & Yelp \\ \hline \hline
BPR-MF & 0.0109 & 0.0076 & 0.0076 & 0.0760 & 0.0147 & 0.0105 & 0.0100 & 0.0119 \\ \hline
GRU4Rec & 0.0116 & 0.0103 & 0.0077 & 0.0104 & 0.0150 & 0.0142 & 0.0097 & 0.0137 \\
Caser & 0.0145 & 0.0114 & 0.0125 & 0.0096 & 0.0226 & 0.0135 & 0.0168 & 0.0129\\
SASRec & 0.0241 & 0.0144 & 0.0245 & 0.0102 & 0.0342 & 0.0177 & 0.0320 & 0.0134 \\ \hline
BERT4Rec & 0.0216 & 0.0143 & 0.0259 & 0.0118 & 0.0300 & 0.0190 & 0.0309 & 0.0171 \\
S\textsuperscript{3}-Rec & 0.0115 & 0.0084 & 0.0314 & 0.0115 & 0.0153 & 0.0111 & 0.0388 & 0.0162 \\
CL4SRec & 0.0268 & 0.0146 & 0.0264 & 0.0131 & 0.0345 & 0.0191 & 0.0339 & 0.0188 \\ \hline
DSSRec & 0.0263 & 0.0139 & 0.0297 & 0.0112 & 0.0329 & 0.0178 & 0.0369 & 0.0152\\
ICLRec & {\ul 0.0326} & {\ul 0.0191} & {\ul 0.0404} & {\ul 0.0153} & {\ul 0.0403} & {\ul 0.0238} & {\ul 0.0480} & {\ul 0.0207} \\
IOCRec & 0.0311 & 0.0169 & 0.0297 & 0.0137 & 0.0396 & 0.0220 & 0.0381 & 0.0192\\ \hline \hline
MCLRec & \textbf{0.0377} & \textbf{0.0201} & \textbf{0.0433} & \textbf{0.0166} & \textbf{0.0455} & \textbf{0.0252} & \textbf{0.0519} & \textbf{0.0220} \\
improv. & 15.64\% & 5.34\% & 7.55\% & 8.63\% & 12.90\% & 5.71\% & 8.06\% & 6.04\% \\ \hline
\end{tabular}
\label{tab:ndcg}
\end{subtable}
\end{table}

\section{EXPERIMENTS}

\subsection{Experimental Setting}

\subsubsection{Datasets.}

We conduct experiments on Amazon~\cite{amazon2,amazon1}, a public dataset of product reviews, and Yelp\footnote{https://www.yelp.com/dataset}, a dataset for business recommendation. In this work, we select three experimental subcategories of Amazon: 'Beauty', 'Toys' and 'Sports'.
Following SASRec~\cite{sasrec}, we retain only the 5-core datasets, where users and items with fewer than five interactions have been removed. Interactions are grouped by user and arranged in ascending chronological order.
For each user, the last item in the interaction sequence is used for testing, the second-to-last item is reserved for validation, and the remaining items are utilized for training. The statistics of three subcategories are displayed in Table \ref{table1}.

\subsubsection{Evaluation Metrics.}

Evaluation involves ranking predictions over the complete set of items without employing negative sampling~\cite{krichene2020sampled,wang2019neural}.
We use two widely-used evaluation metrics to evaluate the model, including Hit Ratio@k (HR@k) and Normalized Discounted Cumulative Gain@k (NDCG@k) where $ k\in\left\{5,10\right\} $.

\subsubsection{Baselines Models.}

We compare our model with baseline methods categorized into four groups. The first group comprises a non-sequential model, BPR-MF~\cite{bpr}. The second group includes traditional sequential recommendation models such as GRU4Rec~\cite{GRU4Rec}, Caser~\cite{caser} and SASRec~\cite{sasrec}. The third group encompasses models that integrate self-supervised learning (SSL) within a sequential framework, including BERT4Rec~\cite{bert4rec}, S\textsuperscript{3}-Rec~\cite{s3rec} and CL4SRec~\cite{cl4srec}. The fourth group consists of intent-based sequential models, namely DSSRec~\cite{dssrec}, ICLRec~\cite{iclrec} and IOCRec~\cite{iocrec}.

\subsubsection{Implementation Details.}

The implementations of Caser, BERT4Rec, S\textsuperscript{3}-Rec, ICLRec and IOCRec are provided by the authors. BPR-MF, GRU4Rec, DSSRec, SASRec and CL4SRec are implemented based on public resources. The parameters for these methods are set as described in their respective papers, and the best settings are selected according to the performance of models on the validation dataset.
We implement our method in PyTorch and use Faiss~\cite{faiss-gpu} for K-means clustering to speed up training. For the encoder part, we set self-attention blocks and attention heads as 2 and embedding dimension as 64. The model is optimized by Adam optimizer~\cite{adam} where learning rate, $\beta_1$ and $\beta_2$ are set to 0.001, 0.9 and 0.999, respectively. The maximum sequence length is set to 50.
For hyper-parameters of MCLRec, we tune $\lambda$, $\beta$ and $\gamma$ all within the set $\left\{0.1, 0.2, ..., 0.8\right\}$. $K$ is tuned in the range of $\left\{32, 64, 128, 256, 512, 1024, 2048\right\}$. The ratio of $R$ to $K$ is set to specific values, with $R/K$ taking on the following proportions: 
$\{0.125, 0.25, 0.375, 0.5, 0.625, 0.75, 0.875\}$.
The model's training incorporates an early stopping mechanism guided by its performance metrics on validation data. All experiments are run on a single NVIDIA A100 GPU.

\subsection{Performance Comparison}

Table \ref{tab:baseline} shows the performance comparison of different methods on four datasets. We have the following observations. 
BPR underperforms compared to conventional sequential models, underscoring the significance of extracting sequential patterns from interaction sequences.
In sequential models, those employing the attention mechanism, such as SASRec and BERT4Rec, outperform models like Caser and GRU4Rec, which do not apply the attention mechanism, demonstrating the attention mechanism's effectiveness in modeling interaction sequences.
Sequential models integrating the attention mechanism, such as SASRec and BERT4Rec, outperform models like Caser and GRU4Rec, which lack this feature. This highlights the effectiveness of the attention mechanism in modeling interaction sequences.
Besides, SSL-based models like BERT4Rec and S\textsuperscript{3}-Rec, which utilize Masked Item Prediction (MIP) tasks to learn user representation, significantly underperform compared to other CL-based models including CL4SRec, ICLRec, IOCRec. The reason might be that MIP tasks for SSL require sufficient context information. 

Intent-based sequential models, including ICLRec, IOCRec and the proposed MCLRec, perform better than SSL-based models, including BERT4Rec and S\textsuperscript{3}-Rec, which indicates the importance of learning intent representations under user-item interaction sequences. However, ICLRec is not as effective as MCLRec, probably because they do not consider the multi-intent level information within one interaction sequence. The performance of MCLRec is also superior to that of IOCRec, which may be attributed to the lack of intersecting multi-intent level information in IOCRec.

Leveraging multi-intent aware contrastive learning, MCLRec demonstrates notably enhanced performance compared to alternative methods across various metrics within three subcategories. The superiority of the best outcome over the top baseline ranges from 2.92\% to 15.64\% in HR and NDCG. 
Low improvement in Sports could be attributed to the increasing of candidate item number and relatively insufficient interaction sequences, which likely introduces greater complexity to contrastive learning tasks in the latent space. 

\begin{figure}[t]
\centering
\includegraphics[scale=0.27]{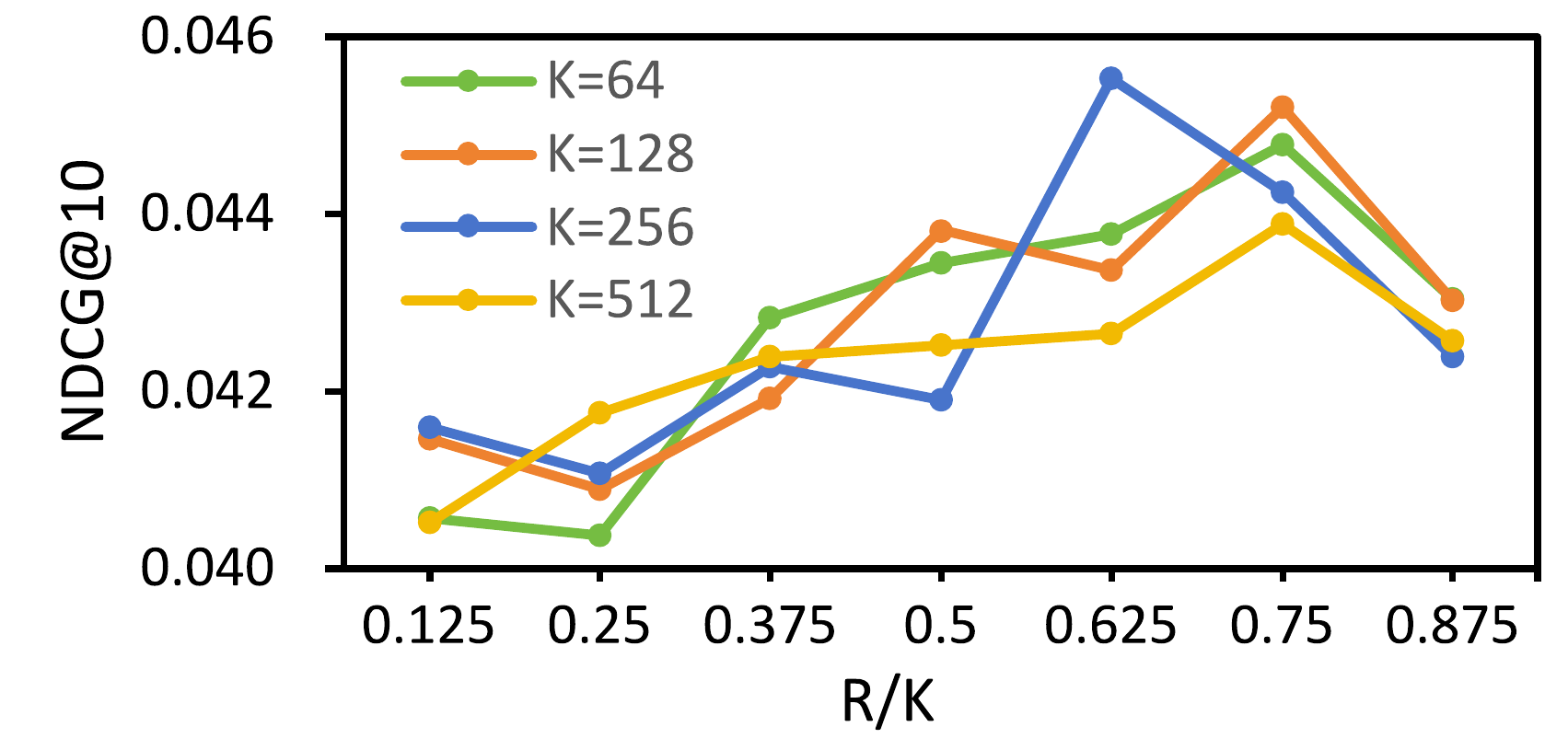}
\caption{Performance comparison w.r.t. hyper-parameters $K$ and $R/K$.}
\label{fig:hyperparameter}
\end{figure}

\subsection{Hyper-parameter Sensitivity}

Given space constraints, we only report the effect of hyper-parameters $R/K$ and $K$ on model performance. We conduct four experiments in the Beauty dataset to investigate the impact of hyper-parameters, including the ratio of relevant intent $R/K$ and the number of intent $K$. We keep other hyper-parameters unchanged for all models in the following experiments and consider NDCG@10. We show the results of experiments in Fig. \ref{fig:hyperparameter} and our observations as follows.

Our method surpasses the baseline across all values of $K$ and $R/K$ (The best NDCG@10 of ICLRec is $0.0403$), which indicates the effectiveness of MCLRec. The model's performance peaks at a ratio of $R/K = 0.625$ given $K = 256$ and $R/K = 0.75$ given $K$ values of $64,128,512$. Subsequently, performance begins to decline as $R/K$ increases further. This decline in performance is likely attributable to introducing an excessive number of representations with low weights $\hat{w}^{u,i}$ into the CL process. This can lead to reduced learning effectiveness and a diminished robustness to noisy data. Additionally, the model's optimal performance improves with increasing values of $K$ up to $256$, beyond which a decline is observed. This suggests that introducing an excessive number of intent categories does not further facilitate multi-intent learning, likely due to the diminished average number of samples available per intent category.

\section{CONCLUSION}

In this paper, we propose a framework called MCLRec, which can filter relevant intents of interactions by similarity of representations in the latent space. We implement a novel multi-intent aware contrastive learning approach to mitigate the impact of the pair-wise representations with high similarity during contrastive learning. Experiments conducted on four public datasets further validate the effectiveness of the proposed model. Several existing studies~\cite{icsrec} have discovered that adopting the perspective of utilizing high-quality positive examples can further enhance the performance and robustness of the SR model, which we leave for future studies.

\section{ACKNOWLEDGEMENT}
This research is supported by the Research Promotion Project of Key Construction Discipline in Guangdong Province (2022ZDJS112) and Shenzhen Technology Univeristy Research Start-up Fund (GDRC202133).

\bibliographystyle{splncs04}
\bibliography{reflist}

\begin{thebibliography}{10}
\providecommand{\url}[1]{\texttt{#1}}
\providecommand{\urlprefix}{URL }
\providecommand{\doi}[1]{https://doi.org/#1}

\bibitem{comirec}
Cen, Y., et.al.: Controllable multi-interest framework for recommendation. In: Proceedings of the 26th ACM SIGKDD International Conference on Knowledge Discovery \& Data Mining. pp. 2942--2951 (2020)

\bibitem{SURGE}
Chang, J., et.al.: Sequential recommendation with graph neural networks. In: Proceedings of the 44th international ACM SIGIR conference on research and development in information retrieval. pp. 378--387 (2021)

\bibitem{SimCLR}
Chen, T., et.al.: A simple framework for contrastive learning of visual representations. In: International conference on machine learning. pp. 1597--1607. PMLR (2020)

\bibitem{IDSR}
Chen, W., et.al.: Improving end-to-end sequential recommendations with intent-aware diversification. In: Proceedings of the 29th ACM International Conference on Information \& Knowledge Management. pp. 175--184 (2020)

\bibitem{iclrec}
Chen, Y., et.al.: Intent contrastive learning for sequential recommendation. In: Proceedings of the ACM Web Conference 2022. pp. 2172--2182 (2022)

\bibitem{bert}
Devlin, J., et.al.: Bert: Pre-training of deep bidirectional transformers for language understanding. arXiv preprint arXiv:1810.04805  (2018)

\bibitem{donkers2017sequential}
Donkers, T., et.al.: Sequential user-based recurrent neural network recommendations. In: Proceedings of the eleventh ACM conference on recommender systems. pp. 152--160 (2017)

\bibitem{fang2020deep}
Fang, H., et.al.: Deep learning for sequential recommendation: Algorithms, influential factors, and evaluations. ACM Transactions on Information Systems (TOIS)  \textbf{39}(1),  1--42 (2020)

\bibitem{MoCo}
He, K., et.al.: Momentum contrast for unsupervised visual representation learning. In: Proceedings of the IEEE/CVF conference on computer vision and pattern recognition. pp. 9729--9738 (2020)

\bibitem{fossil}
He, R., et.al.: Fusing similarity models with markov chains for sparse sequential recommendation. In: 2016 IEEE 16th international conference on data mining (ICDM). pp. 191--200. IEEE (2016)

\bibitem{amazon2}
He, R., et.al.: Ups and downs: Modeling the visual evolution of fashion trends with one-class collaborative filtering. In: proceedings of the 25th international conference on world wide web. pp. 507--517 (2016)

\bibitem{GRU4Rec}
Hidasi, B., et.al.: Session-based recommendations with recurrent neural networks. arXiv preprint arXiv:1511.06939  (2015)

\bibitem{MTAM}
Ji, W., et.al.: Sequential recommender via time-aware attentive memory network. In: Proceedings of the 29th ACM international conference on information \& knowledge management. pp. 565--574 (2020)

\bibitem{cvssl}
Jing, L., et.al.: Self-supervised visual feature learning with deep neural networks: A survey. IEEE transactions on pattern analysis and machine intelligence  \textbf{43}(11),  4037--4058 (2020)

\bibitem{faiss-gpu}
Johnson, J., et.al.: Billion-scale similarity search with {GPUs}. IEEE Transactions on Big Data  \textbf{7}(3),  535--547 (2019)

\bibitem{sasrec}
Kang, W.C., et.al.: Self-attentive sequential recommendation. In: 2018 IEEE international conference on data mining (ICDM). pp. 197--206. IEEE (2018)

\bibitem{adam}
Kingma, D.P., et.al.: Adam: A method for stochastic optimization. arXiv preprint arXiv:1412.6980  (2014)

\bibitem{krichene2020sampled}
Krichene, W., et.al.: On sampled metrics for item recommendation. In: Proceedings of the 26th ACM SIGKDD international conference on knowledge discovery \& data mining. pp. 1748--1757 (2020)

\bibitem{albert}
Lan, Z., et.al.: Albert: A lite bert for self-supervised learning of language representations. arXiv preprint arXiv:1909.11942  (2019)

\bibitem{iocrec}
Li, X., et.al.: Multi-intention oriented contrastive learning for sequential recommendation. In: Proceedings of the Sixteenth ACM International Conference on Web Search and Data Mining. pp. 411--419 (2023)

\bibitem{GeoSAN}
Lian, D., et.al.: Geography-aware sequential location recommendation. In: Proceedings of the 26th ACM SIGKDD international conference on knowledge discovery \& data mining. pp. 2009--2019 (2020)

\bibitem{carnn}
Liu, Q., et.al.: Context-aware sequential recommendation. In: 2016 IEEE 16th International Conference on Data Mining (ICDM). pp. 1053--1058. IEEE (2016)

\bibitem{roberta}
Liu, Y., et.al.: Roberta: A robustly optimized bert pretraining approach. arXiv preprint arXiv:1907.11692  (2019)

\bibitem{coserec}
Liu, Z., et.al.: Contrastive self-supervised sequential recommendation with robust augmentation. arXiv preprint arXiv:2108.06479  (2021)

\bibitem{hgn}
Ma, C., et.al.: Hierarchical gating networks for sequential recommendation. In: Proceedings of the 25th ACM SIGKDD international conference on knowledge discovery \& data mining. pp. 825--833 (2019)

\bibitem{dssrec}
Ma, J., et.al.: Disentangled self-supervision in sequential recommenders. In: Proceedings of the 26th ACM SIGKDD International Conference on Knowledge Discovery \& Data Mining. pp. 483--491 (2020)

\bibitem{amazon1}
McAuley, J., et.al.: Image-based recommendations on styles and substitutes. In: Proceedings of the 38th international ACM SIGIR conference on research and development in information retrieval. pp. 43--52 (2015)

\bibitem{ICM-SR}
Pan, Z., et.al.: An intent-guided collaborative machine for session-based recommendation. In: Proceedings of the 43rd international ACM SIGIR conference on research and development in information retrieval. pp. 1833--1836 (2020)

\bibitem{icsrec}
Qin, X., et.al.: Intent contrastive learning with cross subsequences for sequential recommendation. arXiv preprint arXiv:2310.14318  (2023)

\bibitem{FPMC}
Rendle, S., et.al.: Factorizing personalized markov chains for next-basket recommendation. In: Proceedings of the 19th international conference on World wide web. pp. 811--820 (2010)

\bibitem{bpr}
Rendle, S., et.al.: Bpr: Bayesian personalized ranking from implicit feedback. arXiv preprint arXiv:1205.2618  (2012)

\bibitem{bert4rec}
Sun, F., et.al.: Bert4rec: Sequential recommendation with bidirectional encoder representations from transformer. In: Proceedings of the 28th ACM international conference on information and knowledge management. pp. 1441--1450 (2019)

\bibitem{SINE}
Tan, Q., et.al.: Sparse-interest network for sequential recommendation. In: Proceedings of the 14th ACM international conference on web search and data mining. pp. 598--606 (2021)

\bibitem{caser}
Tang, J., et.al.: Personalized top-n sequential recommendation via convolutional sequence embedding. In: Proceedings of the eleventh ACM international conference on web search and data mining. pp. 565--573 (2018)

\bibitem{ASLI}
Tanjim, M.M., et.al.: Attentive sequential models of latent intent for next item recommendation. In: Proceedings of The Web Conference 2020. pp. 2528--2534 (2020)

\bibitem{MCPRN}
Wang, S., et.al.: Modeling multi-purpose sessions for next-item recommendations via mixture-channel purpose routing networks. In: International Joint Conference on Artificial Intelligence. International Joint Conferences on Artificial Intelligence (2019)

\bibitem{wang2019neural}
Wang, X., et.al.: Neural graph collaborative filtering. In: Proceedings of the 42nd international ACM SIGIR conference on Research and development in Information Retrieval. pp. 165--174 (2019)

\bibitem{RRN}
Wu, C.Y., et.al.: Recurrent recommender networks. In: Proceedings of the tenth ACM international conference on web search and data mining. pp. 495--503 (2017)

\bibitem{SGLRec}
Wu, J., et.al.: Self-supervised graph learning for recommendation. In: Proceedings of the 44th international ACM SIGIR conference on research and development in information retrieval. pp. 726--735 (2021)

\bibitem{dhcn}
Xia, X., et.al.: Self-supervised hypergraph convolutional networks for session-based recommendation. In: Proceedings of the AAAI conference on artificial intelligence. vol.~35, pp. 4503--4511 (2021)

\bibitem{cl4srec}
Xie, X., et.al.: Contrastive learning for sequential recommendation. In: 2022 IEEE 38th international conference on data engineering (ICDE). pp. 1259--1273. IEEE (2022)

\bibitem{sqnsac}
Xin, X., et.al.: Self-supervised reinforcement learning for recommender systems. In: Proceedings of the 43rd International ACM SIGIR conference on research and development in Information Retrieval. pp. 931--940 (2020)

\bibitem{fmfd}
Yao, T., et.al.: Self-supervised learning for deep models in recommendations. arXiv preprint arXiv:2007.12865  (2020)

\bibitem{yao2021self}
Yao, T., et.al.: Self-supervised learning for large-scale item recommendations. In: Proceedings of the 30th ACM International Conference on Information \& Knowledge Management. pp. 4321--4330 (2021)

\bibitem{dream}
Yu, F., et.al.: A dynamic recurrent model for next basket recommendation. In: Proceedings of the 39th International ACM SIGIR conference on Research and Development in Information Retrieval. pp. 729--732 (2016)

\bibitem{MARank}
Yu, L., et.al.: Multi-order attentive ranking model for sequential recommendation. In: Proceedings of the AAAI conference on artificial intelligence. vol.~33, pp. 5709--5716 (2019)

\bibitem{yuan2019simple}
Yuan, F., et.al.: A simple convolutional generative network for next item recommendation. In: Proceedings of the twelfth ACM international conference on web search and data mining. pp. 582--590 (2019)

\bibitem{DeepCoNN}
Zheng, L., et.al.: Joint deep modeling of users and items using reviews for recommendation. In: Proceedings of the tenth ACM international conference on web search and data mining. pp. 425--434 (2017)

\bibitem{s3rec}
Zhou, K., et.al.: S3-rec: Self-supervised learning for sequential recommendation with mutual information maximization. In: Proceedings of the 29th ACM international conference on information \& knowledge management. pp. 1893--1902 (2020)

\end{thebibliography}
\end{document}